# Abiot: A Low cost agile sonic pest control tricopter

Saurabh Kumar Ariyan           Eshant Bagela          Akanksha Priyadarshini  [CS]
3rd year,CS                    2nd year, AEI              2nd year, AEI
 JRE School of Engineering    JRE School of Engineering    JRE School of Engineering

## Abstract

In this paper we introduce the concept of an agile electronic pest control intelligent device for commercial usage and we have evaluated its performance in comparison with other existing similar technologies. The frequency and intensities are changed with respect to the target pest however human behavior has been found to be inert with their exposure. The unit has been tested in lab conditions as well as field testing done have given encouraging results. The device can be a standalone unit and hence work for small scale viz. kitchen garden on the other hand multiple devices acting in coordination with each other give the desired output on a larger scale. Lastly, we chose a tricopter over the wider used quadcopter for greater yaw and agility. The work was funded by IEEE under AiyeHum – 2012 and the prototype was successfully built and tested. The system response effectiveness was found to 86.5% up to a distance of 15 meters.

## I. Problem definition

Insects are found in all types of environment and they occupy little more than two thirds of the known species of animals in the world. Insects affect human beings in a number of ways. Many of them fed on all kinds of plants including crop plants, forest trees, medicinal plants and weeds. They infest the food and other stored products in godowns, bins, storage structures and packages causing huge amount of loss to the stored food and deterioration of food quality. Post harvest loss is estimated to be around 17,000 tons in India alone worth Rs 44,000 crores every year. Insects that cause less than 5 % damage are not considered as pests. The insects which cause damage between 5 - 10% are called minor pests and those that cause damage above 10% are considered as major pests.

The most common method for pest control is the usage of agrochemicals has resulted in heavy amount of pesticides entering the food chain. The potential toxicity of these chemicals to humans and other animals is quite disturbing in light of the Stockholm Convention on Persistent Organic Pollutants (POP) [1], 9 of the 12 most dangerous and persistent organic chemical are pesticide. The World Health Organization reports 220,000 people die every year worldwide because of pesticide poisoning. More so due to lack of infrastructural facilities pesticide residues in food are often higher in developing countries. Farmers who use pesticides have a significantly higher rate of cancer incidence than non-farmers. Thus the solution rates poorly on eco-friendliness.

Electronic pest control [2] [3] has been in use for a long time. It is the name given to the use of any of the several electronic devices designed to repel or eliminate pest. The use of such devices has been mostly for rodents and insects. These devices have no mobility and intelligence for travel and coordination when used in multiplicity. Thus the target area is gets dependent upon the high frequency making it too small for any commercial scale usage but it has been widespread due to its ecological credentials and ease of use.

## II. Solution of the problem defined

The growing concern for eco friendly approaches has made alternative method of pest control increasingly attractive. The problems with the electronic method device have been numerous with many research discarding their effectiveness [4] [5] due to the insufficient data to constant production of high frequency sound. More over apart from the dispute over theoretical claims another challenge remains of restricted area of exposure which makes the device redundant for any commercial usage. The evaluation of proposed solution is done on the basis of (i)Effectiveness in removal of pests from the crops (ii) Minimal usage of chemicals for the same (iii) Price should be in the range of common pesticide to make it feasible for farmers (iv) No special training or other accessibility be needed using the purposed solution

Recent researchers have found veracity in the claims of ultrasonic devices as method of pest control. The extensive study carried by Bhardrirajiu *et al* [7] has found repellent effect of pesticides with a marked reduction in the mating of various insects e.g. crickets. On the other hand the same device had little effect on certain arthropod e.g. cockroaches. We propose a combination of treatments to be explored and innovative methods which have proven effect on a wide range of insects. RF has been used in similar devices earlier [6] and hence we tested extensively a combination of RF and ultrasonic emitter which were found to be working with high efficiency. The variation of frequency range is of significant importance for insects of different species. It has been proven that sufficiently loud sound can kill insects and rodents under laboratory conditions by raising the body temperature. These effects have earlier been disregarded due to expense involved and lack of mobility of the device.[2] [3]

We hence propose a tricopter with attached ultrasonic and RF generator to be used as a solution. The device has exceptional agility inherent to the design which provides ease of motion and faster traversal across the length and breadth of the targeted field. This increase in speed provides more number of loops across the field in same time thus making the exposure time small but repetitive for each area. The aim is to provide a simulation of shocks through each of these traversals. The device is designed with an onboard rechargeable power supply, microcontroller and sensors.

The material should have sufficient strength that continuous exposure to sun is of no significant concern. The best can be carbon fiber for such requirements on the other hand it can largely affect the cost of the system hence we decided to opt for an aluminum frame. An on-board alarm is proposed on indicating the low battery or other system failure causing termination of the run. The system is made to be water resistant to certain degree to make it immune to mild shower and similar possibilities due to field exposure and programmed to land on the same location as the take off. This same location landing is maintained even in case of abnormal termination in case of battery running out or similar mechanical faults. However if at all the system crash-lands at any other place other than the starting place it keeps producing a sound for tacking its present location.

The initial height is programmed to be a height of 1 meter (3 feets) and then it checks the height of crop before any further rise in vertical height. Once the vertical height is found to be optimal i.e. 3 feet or 20 cm above the crop height then only horizontal motion takes place. The outward dimensions of the field are pre feed to the system and the agent provides a radially inward motion until reaching the center of the field. After reaching the center the agent retraces its path. These laps are repeated over the same area for a cumulative effect.

In case of larger fields, many agents can work in coordination with each other. It is suggested that the coordination can be neighbor-to-neighbor basis rather than creating a central server on similar lines as done in rescue operations. Kumar *et al* [8] has quite a remarkable work on path planning for decentralized system and following a similar method has proven to be useful. The major constraint is overlapping or vacant area can occur in multi agent approaches and should be taken care before the path feed is done.

It is suggested to start the agent over the field even before the sowing of seeds for a better results however the effect due to variations of such timing are a matter of considerable study. Around 6 to 7 laps have been found sufficient to keep the pest under check and frequency can be varied for different target pests. It is seen that a decrease in the laps also worked after the crops have been exposed to the agent for a sufficient days. The number of days varies with the crop.

Different studies have attributed very high habituation rate observed in duration as small as ten days however more extensive study is required. The objective has been to keep a broad range of frequency so as different variety of pests can be targeted and thus the circuit for pulse generator is made with a variable capacitance for controlling the change of value of capacitor with associated change in frequency. [Appendix 1]

## III. Results and discussion

The agent has shown an impressive effectiveness of about 83% when used in coordination with other agents while a standalone device can have an effectiveness of upto 89.5%. These upper limits are valid in a range of 10-15m depending upon the crop density and the target pest. These numbers have been generated by keeping the total number of pests of particular specie known before hand and by finding the removed pests afterwards. In comparison with commercial available pesticides these are encouraging results as the efficiency were found to be around 92-93% largely varying by the target pest. However our agent scores over these devices when new specie attacks the region.

The timing of each lap can be reduced by choosing a less dense path. The sound dispersal density on the proposed path is quite high which can be the attributed cause of high efficiency, for one season crop where the fruit is on the apex and the shoot can be compromised we do propose a less intensive path as this can be a direct trade off for increasing the battery life of the agent..

We are already of the view that singularly ultrasonic cannot be used for long term due however such habituation should not be a factor when used in conjunction with RF waves. Empirical proof for the same can be a matter of further discussion.

On the other hand use of tricopter rather than a quadcopter gives the required agility to the system and the speed and thrust provided is greater thus turning the balance in favour of tricopter over the inherent stability provided by the quadcopter.

The results are judged on the basis of no use of harmful chemicals for removal of pests and the development cost is competitive with the leading technology. The end user has only a one time investment with the negligible cost for recharging the batteries. The battery can have an on-board life of approximately one and a half years. Future work may include making the system solar powered or other alternative source and making the agent more disaster resistant. It can also be of relevance to have a more deterministic study of habituation in the pest over prolonged exposure to such devices

## IV. Practical application

The device can affect the pest control in large part with the impressive results and greatly reduce the input price of sharecroppers, the prime target of this work. Apart from apparent use in fields and crop land, the large tech-savvy audience can have a device for their kitchen gardens. Such products are made of carbon fiber for aesthetics purposes as cost do not remain a major consideration.

When the sharecroppers do not have proper knowledge of varieties of pests affecting the crops or in case of a new pest attacking the region the system can be of major use providing a certain kind of food security. Apart from that these can also be used in post harvest scenarios where use of pesticide is of significant concern due to high probability residues contaminating the food chain.

## V. Summary

Our preliminary results of field analysis show the device has potential to eliminate rodents and other common pests. If other finding show similar results the device should be used in filed for an eco friendly pest control and thus saving the food chain from significant damage. The effectiveness is increased due to inherent intelligence in the device in path planning which is unique to each field. The system requires an extensive performance evaluation to determine its efficiency on various pests found in different climatic conditions.